# Be aware of overfitting by hyperparameter optimization!

Igor V. Tetko,*[,1,2] Ruud van Deursen,[3] Guillaume Godin,*[,2]

[1]Institute of Structural Biology, Molecular Targets and Therapeutics Center, Helmholtz Munich - Deutsches Forschungszentrum für Gesundheit und Umwelt (GmbH), 86764, Neuherberg, Germany; [2]BIGCHEM GmbH, Valerystr. 49, 85716, Unterschleißheim, Germany; [3]DSM-Firmenich SA, Rue de la Bergère 7, Satigny, Switzerland

Corresponding authors: Igor V. Tetko, igor.tetko@helmholtz-munich.de and Guillaume Godin, guillaume.godin@gmail.com

## Abstract

- Hyperparameter optimization is very frequently employed in machine learning. However, an optimization of a large space of parameters could result in overfitting of models. In recent studies on solubility prediction the authors collected seven thermodynamic and kinetic solubility datasets from different data sources. They used state-of-the-art graph-based methods and compared models developed for each dataset using different data cleaning protocols and hyperparameter optimization. In our study we showed that hyperparameter optimization did not always result in better models, possibly due to overfitting when using the same statistical measures. Similar results could be calculated using pre-set hyperparameters, reducing the computational effort by around 10,000 times. We also extended the previous analysis by adding a representation learning method based on Natural Language Processing of smiles called Transformer CNN. We show that across all analyzed sets using exactly the same protocol, Transformer CNN provided better results than graph-based methods for 26 out of 28 pairwise comparisons by using only a tiny fraction of time as compared to other methods. Last but not least we stressed the importance of comparing calculation results using exactly the same statistical measures.

**Scientific Contribution** We showed that models with pre-optimized hyperparameters can suffer from overfitting and that using pre-set hyperparameters yields similar performances but four orders faster. Transformer CNN provided significantly higher accuracy compared to other investigated methods.

# Introduction

The water solubility is crucial for different chemistry applications and has been a focus for studies since 1867, when Richardson showed that the toxicities of ethers and alcohols were inversely related to their water solubility.[1] The field is actively developing, and new models and approaches to predict this important property continue to be regularly published.[2–6] Recently, the first EUOS/SLAS challenge for the prediction of solubility classes measured by nephelometry assay was organized on Kaggle.[7] Since the applicability domain of models critically depends on data,[8] the studies reporting new large datasets with solubility data, such as AqSolDB,[9] are of considerable interest to the research community. That is why the recent article by Meng *et al.*,[10] which reported on the collection of large sets of solubility values, while also mentioning a significant drop in the RMSE due to the reported data curation, attracted our attention. One of the important methodological approaches reported in the article was the use of a hyperparameter optimization procedure which required a lot of computational power. Therefore, despite the availability of the authors' scripts, the reproduction of results reported by the authors proved to be very challenging due to the very high computational demands associated with performing the hyperparameter optimization. We were interested in whether similar or better results could be obtained using a moderate number of computational resources (a level of which would typically be available in academic settings, i.e., with several GPU cards) and whether the use of hyperparameter optimisation was really critical for this study.

Therefore in this work, we systematically investigated the impact of the previously reported data processes on calculated results and compared it with the result obtained using relatively modest computational resources, which used pre-optimisation of hyperparameters of machine learning algorithms.

Our main contributions are the following:

- We reinforce the need of careful data cleaning to aggregate multiple sources and avoid data duplication
- We show that hyperparameter optimisation may not provide an advantage over using a set of pre-optimised parameters and may also contribute to overfitting
- We demonstrate that the accuracy of TransformerCNN was higher than that of graph methods ChemProp and AttentiveFP
- We make a clear distinction between the *cuRMSE* (curated RMSE) and the standard *RMSE* (Root Mean Squared Error) function and highlight the importance of using the same statistical measure when comparing results

# Data

The authors collected seven datasets, as summarized in Table 1. In the article, there were three versions of the sets dubbed as "original" ("Org"), "cleaned" ("Cln") and "curated" ("Cure").

## Original sets ("Org")

The "original" sets collected data retrieved by the authors from the respective data sources, as reported in Meng *et al.*[10] Some of the original sets contained duplicates. For example, On-line CHEmical database and Modelling environment (OCHEM)[11] has a policy of collecting data as it is published in the original articles. This policy allows for easier reproduction of the respective studies. Thus, if some data were repeatedly reused in other publications, many of the records would likely be duplicated. The original Huuskonen data set[12] and its curated version[8] (AQUA set in Table 1) were re-used in practically all publications on the prediction of solubility and made up major parts of the PHYSP, ESOL[13] and OCHEM sets in particular. Moreover, since the data in this set were extracted from the AqSolDB database,[9] they were also present in the AQUASOL set. The AQUASOL set, as any other set in this article, is a collection of data from multiple sources. It also contained a number of single heavy atom molecules, such as [CH4], [Mg2+], [Mo], [Re], [H+].[F-] etc., or inorganic complexes [O-2].[O-2].[Mg+2].[Ca+2], [Al+3].[Cl-].[Cl-].[Cl-], etc., which could not be processed by graph-based neural networks due to the fact that there are no bonds between heavy atoms set for those objects to apply graph convolution or/and these atom/molecule/compound types are not supported. The authors removed duplicates and metals during the cleaning and standardization procedure to create "clean" and "curated" sets as described below, and the 192 metal-containing compounds that remained in the "Org" set were excluded as reported in Table 1.

## Clean sets ("Cln")

The cleaning procedure (described in Meng *et al.*[10]) included SMILES standardization using MolVS followed by the removal of duplicates (only when the difference between values for the same molecules in two records was less than 0.01 log unit), removing records that followed non-standard experimental protocols (temperature 25 ± 5°C, pH 7 ± 1) as well as the removal of compounds containing metals.

OCHEM uses InChi keys to index molecules as well as rounded (0.01 log unit) property values to calculate unique keys. Using such keys, we identified and eliminated a few additional exact duplicates for "Cln" thermodynamic solubility sets, including molecules containing metals, which were initially not detected by the authors (Table 1).

The data for kinetic solubility ("KINECT" dataset) were all downloaded from the OCHEM database. However, it appears as though the authors did not initially notice that the majority of these records in OCHEM originally came from the PubChem AID1996 assay,[14] which contained 57858 measurements. OCHEM contains the original data uploaded from this assay as well as data from two articles[15,16] that used this assay in their studies. These three sources contributed a total number of 161710 records out of 164273 records in the "KINECT Org" set. The different processing procedures (salt elimination, neutralization, aromatisation, etc.) in the aforementioned studies,[15,16] resulted in different chemical structures. Hence, after the deduplication procedure Meng et al[10] obtained 82057 records instead of the original 57858, i.e., there were 24199 duplicated measurements. Examples of duplicated structures are shown in

Figure 1. Some duplicates were SMILES with and without reported stereochemistry, ionized and neutral compounds, etc. which appeared due to different data standardization procedures in the respective data sources. Such a high value of duplicates (>37%) could imply a biased estimation accuracy of the developed models for kinetic solubility.

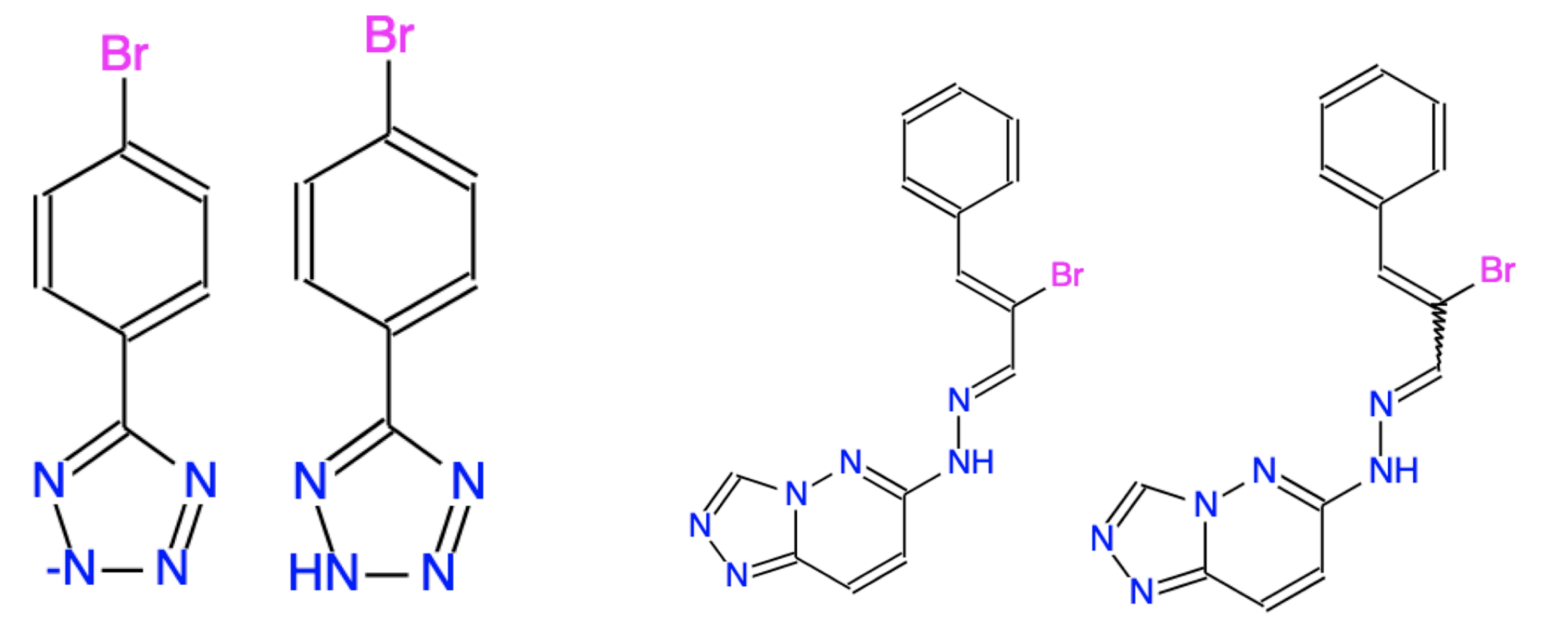

LogS=-4.37 LogS=-5.14

Brc1ccc(-c2nn[n-]n2)cc1 Br/C(C=NNc1ccc2nncn2n1)=C\c1ccccc1

Brc1ccc(-c2nn[nH]n2)cc1 BrC(=C\c1ccccc1)/C=N/Nc1ccc2nncn2n1

**Figure 1.** Examples of duplicated structures with exactly the same solubility values but different SMILES in the kinetic solubility dataset. These and other similar duplicated structures contributed about 37% of records in the combined kinetic solubility set of Meng et al.[10]

For the CHEMBL set, 30099 values were reported in the article[10] but 30099 and 31099 records were made available in the repository for “Org” and “Cln” sets, respectively. For the “Cln” set, we used 31099 records as provided in the article’s repository (the chembl_stand_clean.csv file). The differences in the number of molecules in the datasets could contribute some changes in calculated statistical parameters.

In the case of duplicated records for analyzed molecules within the same set, the authors used a weighting to avoid their overrepresentation during learning (so-called “inter-dataset curation”). The authors assigned each record a weight, inversely proportional to the total number of records per molecule. Thus each molecule had a total weight equal to “1”.

Table 1. Analysis and cleaning of clean data (“Cln”) reported in the study by Meng *et al.*[10]

| Dataset | Number of records | | | Remark to OCHEM cleaning (see also | Unique |
|---|---|---|---|---|---|
| | “Org” | Initial | After | | |

| | set” | “clean” records in ref[10] | OCHEM cleaning | text) | stereochemical molecules (or ignoring stereochemistry) |
|---|---|---|---|---|---|
| AQUA | 1311 | 1311 | 1310 | 1311 were reported in the article[10] but 1310 values were used as reported in the GitHub repository | 1301 (1301) |
| PHYSP | 2010 | 2001 | 2001 | | 2001 (2001) |
| ESOL | 1128 | 1116 | 1115 | 1 duplicate was eliminated | 1109 (1108) |
| OCHEM | 6525 | 4218 | 4177 | 41 duplicates were eliminated | 3620 (3568) |
| AQSOL | 9982 (9790)[1] | 8701 | 8687 | Six duplicates and eight molecules with metals were removed from “clean” set | 8674 (8394) |
| CHEMBL | 30099 | 30099 | 31050 | 30099 were reported in article, but 31099 values were found and used from the GitHub; 31 non-organic compounds and 18 duplicates were removed | 26377 (25796) |
| KINECT | 164273 | 82057 | 60392 | Multiple duplicated data from the same assay were removed, see section “Data” | 60233 (60233) |

[1]192 metals and non-organics were removed from the “Org” AQSOL set.

## Curated sets (“Cure”)

The authors (Meng *et al.*) also produced curated datasets. The authors first assigned weights to datasets corresponding to their quality (high quality: AQUA, PHYSP, ESOL with weight “1.0” and OCHEM with weight 0.85 as well as low quality: AQSOL with weight 0.4 and CHEMBL with weight 0.8), which was determined manually based on the performance of ChemProp. In the first step of this analysis, records were assigned a weight based on the sets they originated from. After that, the authors extended each analyzed set with records for the same molecule found in other sets. In cases where several solubility values were present for the same molecule, and the differences between their values were less than d=0.5 log units (corresponding to the estimated experimental accuracy of solubility measurements[8]), their values were merged and the weight of the merged record was updated. Otherwise, records were kept with the weights assigned according to their respective datasets. For more details about this procedure, we refer to the original article by Meng *et al.*[10] As a result of this procedure, new solubility sets with weights for each record were obtained and provided by the authors, which were named “curated” (Cure) datasets.

Table 2 indicates the number of records, the average weight for all records in the respective dataset, and the performance of the models using clean and curated data. We noticed that results reported for the PHYSP set in ref[10] are inconsistent and likely to be a reporting error.

Table 2. Summary of clean and curated sets and model performance based on these sets extracted from ref.[10] and its GitHub[17]

| Dataset | Clean (“Cln”) dataset | | | | Curated “Cure” dataset | | | |
|---|---|---|---|---|---|---|---|---|
| | records | average weight | Published cuRMSE | | records | average weight | Published cuRMSE | |
| | | | ChemProp | AttFP | | | ChemProp | AttFP |
| AQUA | 1311 | 0.993 | 0.58 ± 0.06 | 0.64 ± 0.01 | 1354 | 0.866 | 0.54 ± 0.04 | 0.58 ± 0.02 |
| PHYSP | 2001 | 1 | 0.60 ± 0.03* | 0.64 ± 0.01* | 2001 | 1 | 0.52 ± 0.02* | 0.55 ± 0.01* |
| ESOL | 1116 | 0.995 | 0.62 ± 0.04 | 0.64 ± 0.03 | 1157 | 0.866 | 0.51 ± 0.05 | 0.59 ± 0.02 |
| OCHEM | 4218 | 0.869 | 0.64 ± 0.04 | 0.65 ± 0.02 | 3766 | 0.712 | 0.52 ± 0.02 | 0.60 ± 0.01 |
| AQSOL | 8701 | 1 | 0.82 ± 0.04 | 0.76 ± 0.01 | 9061 | 0.496 | 0.52 ± 0.01 | 0.59 ± 0.01 |
| CHEMBL | 30099 | 0.848 | 0.81 ± 0.02 | n/a | 28675 | 0.325 | 0.50 ± 0.01 | n.a. |
| KINECT | 82057 | 1 | 0.431 ± 0.003 | n/a | 81935 | 0.999 | 0.43 ± 0.003 | n.a. |

*The clean and curated sets results for the PHYSP set are inconsistent and some of them are likely to be a reporting error.

## Statistical parameters

One of the traditional statistical parameters used to estimate the accuracy of models is RMSE (eq 1)

$$RMSE = \sqrt{\sum_{i=0}^{n-1} \left(\overline{y_i} - y_i\right)^2 / n} \quad (1)$$

However, the authors used a modified RMSE (an *ad hoc* “curated RMSE” or “cuRMSE”), which incorporated the weights of records to estimate the performance of their models (see eq 2)

$$cuRMSE = \sqrt{\sum_{i=0}^{n-1} w_i * \left(\overline{y_i} - y_i\right)^2 / n} \quad (2)$$

In both formulas $\overline{y_i}$ are predicted values (which were averaged across several models to improve model accuracy, as stated by Weng *et al.*[10]) and $y_i$ are experimental values.

The cuRMSE is an interesting loss function for training of neural networks, since it decreases the impact of some individual data points. However, this measure depends on the distribution of weights and could potentially be a source of bias when comparing model performance across different sets. Indeed, it decreases errors for molecules that have more than one value in the dataset. For example, if a molecule has two records, each with the same difference between predicted and experimental values, e.g., 0.6, and each record has the same weight 1/2 = 0.5 for all records (as seen in "Cln" datasets to account for inter-set redundancy), its cuRMSE will be sqrt((0.5*0.6*0.6 + 0.5*0.6*0.6)/2)=0.3, i.e., its cuRMSE will be artificially halved in comparison to the RMSE for the same molecule. Assignment of small weights to records from datasets with high errors may have an even more pronounced effect.

# Methods

In this section, we examine whether or not hyperparameter optimization provides a significant improvement in the performance of analyzed methods.

## Analyzed methods

Attentive FingerPrint (AttFP),[2] ChemProp[18] and Transformer CNN[19] methods were used for the analysis. The first two methods were based on keras GCNN (KGCNN) repository code.[20] While a ChemProp implementation was available, we preferred to use the KGCNN implementation since it allowed for better control as well as optimization of several hyperparameters by testing the algorithm against datasets from our previous study.[21] The implementation of Transformer CNN is available elsewhere.[22] All three methods are available as part of openOCHEM software.[23] Both AttFP and ChemProp are generally among the Top 5 best graph models applied to physical property predictions, including 3D graphs. However, because both methods are based on RDKit, there were some cases of failure on compounds that could not be processed by this package or by each method. The Transformer CNN is a non-graph method based on the Transformer architecture,[24] which analyzes the representation of compounds as SMILES strings. This method was added to provide a point of comparison with graph-based architectures. All results from this study are publicly available online[25] and instructions on how to reproduce the calculations (and/or assess development models) are provided on GitHub.[26]

Meng *et al.*[10] performed an extensive optimization of hyperparameters using a large GPU cluster. In particular, the authors noted that one round of calculations (solely for results of the ChemProp method) reported in Table 3 of their article (half of which are listed in Table 3 below) took approximately two weeks on a cluster with 1200 compute nodes (38,200 cores and 4800 GPU accelerators). Thus, the process required >1.8M (4800*24*16) hours of calculations on GPU cards. Moreover, the authors stated that they could not produce results with AttFP for the CHEMBL and KINECT sets since the calculations would be too time-consuming. Since we did not have access to such a powerful cluster, we decided to skip parameter optimization and use default hyperparameters provided by OCHEM developers for the respective methods and implemented in openOCHEM.[23] The employed hyperparameters were selected based on analysis of several small sets used in our previous study.[21] We performed our analysis on a communal cluster with 16 GPU cards, usually running two tasks simultaneously as it was typically faster than running only one task per card. The training of all three models in Table 3 typically required less than 6 hours on all available cards (ca 100 GPU hours), with the exception of the largest “KINECT Org” set which required about 100 GPU hours for the development of models with three analyzed methods. Assuming that the GPU cards we used (Nvidia GeForce RTX 2070, Titan V, Tesla V100) had similar specifications to those used by Meng *et al.*[10] (the authors did not specify which cards were used for production, but we assume standard hardware accessible on cloud services for this comparison, with a speed similar to or higher than that of cards used by us), all results reported in this study would require about 10,000 times less computational power, i.e., less than 3 minutes of calculations on their GPU cluster.

The workflow for molecule processing in OCHEM included standardization, de-salting (keeping the major fragment), and neutralization. We noticed that graph-based methods failed for some compounds in some sets (e.g., AQSOL, CHEMBL, KINECT) because of an RDKit error after the neutralization of molecules. For smaller sets, we noticed that both results of the methods were similar, both with and without neutralization. Therefore for these two methods, the neutralization step was skipped. The partitioning of data on folds (see below) was the same for all three analyzed methods, which allowed direct comparison of results of models developed with different algorithms.

## Validation protocols

The authors Meng *et al*[10] used random and scaffold partitioning [0.8,0.1,0.1] for training, testing and evaluation, which was repeated five times. We decided to analyze results obtained with random partitioning, since the issues we encountered are also relevant to partitioning based on scaffold splitting. The protocol employed by the authors was very similar to 10-fold cross-validation (10-fold CV), which was used in this study for all results reported in openOCHEM. Indeed, in both protocols, 10% of data were excluded from model training and these data were predicted once hyperparameter optimization was finished. OCHEM 10-fold CV provides prediction for all the data, as in any CV approach. The internal procedure in OCHEM splits data into internal training (81%), early stopping (9%) and evaluation sets (10%) corresponding to [0.81,0.09,0.1] split, which is practically the same as the [0.8,0.1,0.1] split used by the authors.

The authors, however, did not run 10-fold CV to obtain predictions for all data, opting instead for five random data splits. Because of the random split, less than 50% of the data was used to calculate the reported model performance in the evaluation set. This procedure should provide very similar results to the 10-fold CV. Also, the authors generated an ensemble of eight models for each split and took their average to improve their results. The apparent disadvantage of the authors' procedure, besides the fact that it does not predict all data, is that it is more computationally expensive, i.e. 5 x 8 = 40 models were developed, while just 10 models were required in OCHEM. The eight models that were used to calculate confidence intervals which, in our opinion, come at too high a computational price compared to simple bootstrap procedure used in OCHEM.[27] OCHEM splits data using the non-stereochemical part of the InChi hash key (to ignore stereochemistry) rather than with canonical SMILES, which was used by the authors of Meng *et al*[10]. The OCHEM bootstrap procedure is more reliable since its splits are insensitive to possible errors related to the stereochemistry of the analyzed molecules.

However, the validation procedures used in Meng *et al.*[10] and this study are very similar and, importantly, both use 90% of data to develop models to predict the respective validation sets used to estimate model performance. This allows for a direct comparison between the results of this and the prior study.

# Results

## Analysis of performance of models for "Org" sets

In order to investigate the effect of hyperparameter optimization and ensemble averaging used by the authors of Meng *et al.*,[10] we compared our results to their published results using 10-fold CV for the "Org" sets (Table 3). The bootstrap procedure estimated the confidence intervals for results reported in this study, as described elsewhere.[27] This procedure, in general, provided smaller confidence intervals, which could be attributed to the fact that it used data from the whole set compared to the analysis in ref[10] which used about 50% of data. The additional variance in the results of Meng *et al.*[10] could be due to hyperparameter selection. The confidence intervals for both analyses decreased with the dataset size.

Table 3. Comparison of RMSE for analyzed original ("Org") datasets.

| Dataset | Published results[10] (following hyperparameters optimisation) | | Results from this study (using default hyperparameters of methods) | | |
|---|---|---|---|---|---|
| | | | RMSE, validation by molecule | | |
| | ChemProp[c] | AttFP | ChemProp | AttFP | Transformer CNN |
| AQUA | 0.58 ± 0.04 | 0.62 ± 0.03 | 0.58 ± 0.02 | **0.60 ± 0.02** | 0.56 ± 0.02* |
| PHYSP | **0.55 ± 0.03[a]** | 0.65 ± 0.02 | 0.57 ± 0.01 | **0.59 ± 0.01** | 0.56 ± 0.01* |
| ESOL | **0.60 ± 0.08** | 0.64 ± 0.02 | 0.61 ± 0.01 | **0.59 ± 0.02** | 0.58 ± 0.02* |
| OCHEM | **0.55 ± 0.02** | **0.60 ± 0.01** | 0.63 ± 0.01 | 0.65 ± 0.01 | 0.59 ± 0.01** |
| AQSOL[b] | 1.02 ± 0.04 | **0.83 ± 0.03** | **1. ± 0.02** | 1.01 ± 0.02 | 0.99 ± 0.01* |
| CHEMBL | 0.92 ± 0.02 | n/a | **0.88 ± 0.01** | 0.93 ± 0.01 | 0.86 ± 0.01** |
| KINECT | **0.401 ± 0.001** | n/a | 0.434 ± 0.002 | 0.465 ± 0.001 | 0.41 ± 0.002** |

[a]Smaller RMSE errors for pairwise comparison of values obtained with the same method in this and the previous study (e.g., calculated AttFP RMSE for AQUA set was smaller, i.e. 0.60 vs 0.62, in this study) are highlighted in bold. [b]149 records for metals or metallo-complexes failed with both ChemProp and AttFP, therefore they were excluded from the set. *Star indicates that Transformer CNN had lower RMSE compared to both ChemProp and AttFP models developed using exactly the same protocol in openOCHEM. **Two stars indicate cases for which Transformer CNN yielded a significantly lower error ($p<0.05$, according to *t-test*) compared to both ChemProp and AttFP models developed using openOCHEM. All values were rounded to one significant digit, which is the default setting in the OCHEM.

The RMSE values calculated with models developed by the authors and recalculated in this article are generally similar. In five cases, the model developed with default parameters in this study provided lower RMSE than those following hyperparameter optimization and in six cases an opposite result was found. In only one case (AttFP model developed for AQSOL dataset), hyperparameter optimization provided a significantly lower RMSE than the value obtained in this study. The opposite was seen during the analysis of the CHEMBL dataset with the ChemProp model, for which the Transformer CNN model yielded a significantly lower RMSE. Based on these observations, we propose that hyperparameter optimization possibly led to overfitting to the data, thus resulting in worse performance, particularly for smaller sets.

Therefore, one could question the need to perform extensive hyperparameter optimization. Indeed, only in one case did the use of hyperparameter optimization lead to improved model performance and this result required 1.8M hours of GPU cards and two weeks of calculations on an HPC cluster.

The Transformer CNN generally provided higher-accuracy predictions than those provided by ChemProp and AttFP in 13 out of 14 comparisons performed in this study; in only one case did it give a lower performance when using the same data and validation folds.

## Analysis of results for “CIn” dataset

The following analysis was performed to evaluate the effect of intra-set curation on the performance of developed models.

cuRMSE values were reported by the authors in their article for the “CIn” and “Cure” sets. As mentioned, this measure may not allow a faithful comparison of model performances across different sets. Since weights were available for both “CIn” and “Cure” sets, we used them to calculate cuRMSE values to allow us to compare models from this study with those previously published by the authors. To this end, we first developed models using the default 10-fold CV procedure of openOCHEM (split by molecule) using “CIn” data. Then, values provided by the cross-validation procedure were used to calculate cuRMSE using eq (2). We excluded the KINECT set from this analysis since it had a different number of records after our curation, but still included its results using openOCHEM for model comparison.

The cuRMSE was not significantly different from RMSE for all "Cln" sets, with the exception of OCHEM and CHEMBL, which had the largest numbers of molecules with several measurements in the dataset.

Performances of methods developed in this study without hyperparameter optimization provided lower cuRMSE in 7 of 12 pairwise analyses (Table 4). Only AQSOL results using AttFP calculated by Meng *et al.*[10] had a lower cuRMSE than the results of this study.

Table 4. Comparison of results for "Cln" dataset.

| Dataset | RMSE | | | cuRMSE | | | |
|---|---|---|---|---|---|---|---|
| | Calculated in this article using fixed set of hyperparameters | | | Same as in first two columns converted to cuRMSE | | Two last columns are published results[10] (following hyperparameters optimisation) | |
| | ChemProp | AttFP | Transformer CNN | ChemProp | AttFP | ChemProp | AttFP |
| AQUA | 0.56 ± 0.02 | 0.57 ± 0.02 | 0.56 ± 0.02 | **0.56± 0.02**[a] | **0.57 ± 0.02** | 0.58 ± 0.06 | 0.64 ± 0.01 |
| PHYSP | 0.57 ± 0.01 | 0.59 ± 0.01 | 0.56 ± 0.01* | **0.57 ± 0.01** | **0.59 ± 0.01** | 0.60 ± 0.03 | 0.64 ± 0.01 |
| ESOL | 0.62 ± 0.02 | 0.62 ± 0.02 | 0.58 ± 0.02** | 0.62 ± 0.02 | **0.62 ± 0.01** | 0.62 ± 0.04 | 0.64 ± 0.03 |
| OCHEM | 0.67 ± 0.01 | 0.69 ± 0.01 | 0.65 ± 0.01** | 0.64 ± 0.01 | 0.65 ± 0.01 | 0.64 ± 0.04 | 0.65 ± 0.02 |
| AQSOL | 0.81 ± 0.01 | 0.82 ± 0.01 | 0.79 ± 0.01** | **0.81 ± 0.01** | 0.82 ± 0.01 | 0.82 ± 0.04 | **0.76 ± 0.01** |
| CHEMBL | 0.84 ± 0.01 | 0.90 ± 0.01 | 0.83 ± 0.01* | **0.72 ± 0.01** | 0.78 ± 0.01 | 0.81 ± 0.02 | n/a |
| KINECT | 0.408 ± 0.002 | 0.443 ± 0.002 | 0.410 ± 0.002 | | | | |

a) Smaller RMSE errors for the pairwise comparison of values obtained for the same method using the same measure, cuRMSE, are highlighted in bold (see also explanations in Table 3). *Star indicates sets for which Transformer CNN yielded a lower error compared to both ChemProp and AttFP models developed using openOCHEM. **Two stars indicate cases for which Transformer CNN yielded a significantly lower error ($p<0.05$, according to t-test) compared to both ChemProp and AttFP models developed using openOCHEM. The values were rounded to one significant digit, which is the default setting in OCHEM.

The use of Transformer CNN gave rise to the lowest RMSE for five out of seven analyzed sets while for the two other sets, the RMSE was more or less equal between Graph and NLP methods. Thus, our main finding when using cuRMSE is that the approach used by the authors (curation of records, use of hyperparameter optimization and ensembling) provided similar results compared to the training of models with pre-optimized hyperparameters without any data weighting. This is an important result since pre-optimization of hyperparameters using small sets could dramatically decrease computational costs.

## Analysis of results for “Cure” dataset

In the final analysis, we compared the effect of intraset data curation on model performance. For this analysis, we reused cross-validation results from the “CIn” set and re-calculated cuRMSE using weights and experimental values provided by the authors (predictions for the weighted set were taken from 10-fold CV and molecules in both sets were matched using InChi).

Table 5. Comparison of cuRMSE for “Cure” datasets.

| | Published results from ref.[10] | | Results from Table 4 re-calculated using cuRMSE | |
|---|---|---|---|---|
| dataset | ChemProp | AttFP | ChemProp | AttFP |
| AQUA | 0.54 ± 0.04 | 0.58 ± 0.02 | **0.50 ± 0.02**[a] | **0.52 ± 0.02** |
| PHYSP* | 0.52 ± 0.02 | 0.55 ± 0.01 | 0.57 ± 0.01 | 0.59 ± 0.01 |
| ESOL | **0.51 ± 0.05** | 0.59 ± 0.02 | 0.54 ± 0.02 | **0.54 ± 0.02** |
| OCHEM | **0.52 ± 0.02** | 0.60 ± 0.01 | 0.54 ± 0.01 | **0.55 ± 0.01** |
| AQSOL | 0.52 ± 0.01 | 0.59 ± 0.01 | 0.52 ± 0.01 | **0.53 ± 0.01** |
| CHEMBL | 0.50 ± 0.01 | n.a. | **0.45 ± 0.01** | 0.48 ± 0.01 |

a) Smaller RMSE errors for pairwise comparison of values obtained for the same method are highlighted in bold. *Results for the PHYSP set were excluded from the analysis due to possible errors in the data curation procedure and reported values.

Only the PHYSP set results calculated in this study had higher RMSE than those reported by Meng *et al.*[10] (Table 5). However, as previously mentioned, the PHYSP set for “CIn” and “Cure” studies was exactly the same, so it could be that results reported for “Cure” are biased.

If we exclude this set, results calculated in this study with default hyperparameters had lower cuRMSE in 6 out of 9 cases compared to those reported by the authors. These results confirm our previous finding that models that have undergone hyperparameter optimization did not yield better results than models using a fixed set of pre-optimised hyperparameters, as investigated in this study.

# Discussion

As we were impressed by the high performance of models reported by the authors[10], we sought to investigate their proposed methodology in-depth and hoped to reproduce their results. However, since we did not have access to the same level of computational power as the authors, we analyzed the results calculated without hyperparameter optimization.

First, our analysis of the original sets ("Org") showed that the methods used in this study gave rise to similar RMSE values to those reported by the authors, despite their extensive hyperparameter optimization.

Moreover, another analysis of results from the "CIn" set (results for "Cure" were obtained from those for "CIn") also showed that hyperparameter optimization did not provide consistent improvement compared to the use of a fixed set of pre-optimized hyperparameters. While hyperparameter optimization is recommended for some of algorithms used in this study, the availability of a cluster with hundreds of GPU cards is a luxury rather than a typical situation for researchers and the use of such clusters would go against the Green AI principles.[28] Indeed, all results performed in this study required >10,000 times fewer computational resources used by Meng *et al.*[10] and even provided better performance in 18 of 34 pairwise comparisons. The lower performance of models after hyperparameter optimization could result from overfitting[29] by hyperparameter selection. This problem is frequently underestimated but should be carefully addressed, especially when using heavy hyperparameter selection for small sets.

Although the hyperparameter optimization method is appealing, its usefulness may be limited depending on the application. In particular, for many important biological (e.g, blood-brain barrier, toxicity, ready biodegradability) or physico-chemical properties (e.g., odor threshold), chemical datasets tend to be composed of a few hundred to at most 1000 data points. For such small datasets, we typically observe strong performance fluctuations between data splits. Thus, relying on one particular split can result in the selection of hyperparameters that are optimal for a specific split but not in general. To more fairly compare results of hyperparameter optimization, a full n-fold would need to be computed, which further increases the computational cost of hyperparameter optimization. This should be done ideally with a repetition of n-fold splits to get statistically significant results. From these repeated experiments one may actually observe that there is no single best solution, but rather a set of conditions defining the set of best available options. These sets of conditions may then be used to design an ensemble model to obtain the greatest possible generalizability. For larger datasets, the split fluctuations are frequently within statistical error and typically representative of any other split (see Table 3 c : standard deviation decreasing with dataset size increasing), but this is where the cost of hyperparameter optimization explodes and makes this approach computationally expensive. More generally, we observed only a few marginal impacts of hyperparameter optimization on the RMSE performance.

Typically, once methods like hyperparameter optimization become a standard feature in various commercial and open-source toolkits,[30] they tend to be used blindly. We encourage the research community to benchmark their models via robust, automated protocols such as OCHEM to receive an unbiased assessment about their performances with smaller sets before starting to use them for more expensive experiments as described in Meng *et al.*[10]. We have shown that data augmentation can provide a performance boost to models developed with limited data, like Transformer CNN models. Other alternatives, such as Gaussian Process and Normalizing Flows models could be evaluated in future studies.[31–33]

Of course, hyperparameter optimization is an important part of model development. However, the readers should be aware that selection of hyperparameters can contribute to overfitting and that the result of hyperparameters selection should not be tested on the same set used for hyperparameters selection. The procedure proposed in our previous study[34] is fully applicable to prevent overfitting by hyperparameter optimization. We suggested splitting the model development process (which can include variable selection, hyperparameter optimisation, etc.) into two steps:

1) "*Model development. Develop your model using your favored method(s) and all available data. Once this step is completed, estimate the accuracy of your final model as follows.*

2) *Model validation.*
    a) *Divide your initial set into n-subsets (e.g., n = 5 was used in this study, larger n or LOO can be recommended for small data sets).*
    b) *Select one subset as the validation set.*
    c) *Use the remaining n−1 sets to develop a model using* ***exactly the same approach*** *as in step 1.*
    d) *Apply the model to the validation set and store the predictions.*
    e) *Return to step b) and repeat the analysis until all subsets are used as the validation sets.*
    f) *Estimate the performance of your model using values calculated from step d).*"[34]

This procedure is part of the OCHEM workflow.

Overfitting is a process in which a model starts to learn noise in the data and, as result, predicts new data, which were not used for model development, with a lower accuracy than a correctly trained model would. It should be mentioned that, in general, overfitting is difficult to prove. Indeed, in addition to overfitting there is also a closely related applicability domain issue. Thus, if data comes from a different distribution and lies outside of the applicability domain, the model performance will also deteriorate. Overfitting happens when data in the test set all come from exactly the same distribution. Very importantly, these data should not be used for model development, since the researchers would obtain overfitted results and proof of overfitting would be difficult to identify. In this respect, we should clearly state that the cross-validation procedure used by Meng et al,[10] was performed correctly: the authors did not use data from the respective test folds to tune their model hyperparameters. Therefore, by comparing their 10-fold CV results to our own, we can demonstrate that their procedure resulted in overfitting. Had they reported overfitted results, one would need to use external sets to prove the effect of overfitting.

Model overfitting can be clearly observed when analyzing the results of challenges, in which the organizers explicitly reported that the data for leaderboard and test sets were randomly split. For example, the organizers of the EUOS/SLAS solubility challenge noticed that a "higher performance on the public leaderboard was most likely due to its exposure to multiple submissions by each team, leading to an overfitting".[35] The same observation was also made during the ToxCast challenge[36,37] to predict Lowest Effect Level toxicity of chemicals, organized by the US Environmental Protection Agency. In this challenge, the leaderboard set scores calculated according to eq. 3 (n=80 compounds, reported as “final provisional score” in the referenced document) where all higher than the scores for blind set (n=63, “system score”).[36]

$$Score = 1000000 * (2 - RMSE) \tag{3}$$

where RMSE was given by eq (1).

We also observed a similar effect on the differences between the methods’ performances for the leaderboard and blind test set during the Tox24 challenge.[38] Moreover there was also a correlation (R=0.36) between the model’s performance difference ($\Delta RMSE = RMSE_{blind} - RMSE_{leaderboard}$) and the number of model submissions: participants that submitted a large number of models tended to obtain larger $\Delta RMSE$ values, thus their estimations of model performance on the leaderboard set became more overfitted and did not reflect the actual performance of the models for the blind set.

Thus, we can conclude that the problem of overfitting by model optimization is an important issue which should be properly addressed by users.

It should be mentioned that the hyperparameters of ChemProp, AttFP, and TransformerCNN were optimized using ten datasets from our previous study.[21] This optimization was performed when these methods were added to OCHEM (in 2019 for Transformer CNN and in 2021 for the other two methods) and was not in any way related to this study. The selected hyperparameters were used as default hyperparameters within OCHEM in several publications, e.g., see[5,39] including the SLAS challenge.[7]

In addition to ChemProp and AttFP models, we reported results from a Transformer CNN model, which in 26 of 28 cases provided a lower RMSE compared to both graph-based methods using the same cross-validation procedure and the exact same data splits (Table 3 & 4). Moreover, in 18 of these cases, Transformer CNN provided significantly lower RMSEs ($p<0.05$). The same approach provided the highest individual score among 30 analyzed models in the Kaggle First EUOS/SLAS Joint Compound Solubility Challenge.[7]

The use of curated procedures employed by authors Meng *et al.*[10] in most cases provided a similar or lower performance compared to the use of datasets without any weighting for inter- and intra-set data curation procedures when using cross-validation. Thus additional studies may need to be provided to confirm the impact of the procedures proposed by the authors. Indeed, the authors provided a weighting according to the data source. However the records in each set were collected from multiple sources with different experimental accuracies.Therefore, the data quality in the merged sets can be markedly varied, and assigning the same weight to all data points in the dataset may be insufficient. The procedure used by the authors could be interesting to compare with an automatic weighted statistical average from multiple sources exhibiting experimental drift to reduce the data variance between sources proposed elsewhere,[40] which may be more appropriate for these data.

We also warn that cuRMSE (which is also dependent on the weights of records) is generally not comparable to RMSE. In Meng *et al.*[10], the comparison of cuRMSE and RMSE within the same tables gave the impression that data curation decreased errors, which was not the case when exactly the same measure, RMSE or cuRMSE using the same weighting, were used for comparison.

We have also identified that the data curation procedures from refs,[15,16] applied to the PubChem AID1996 assay[14] data resulted in 24199 duplicated records, which had either different structures or rounded values and thus could have been treated as new data by Meng *et al.*[10] Similar data duplication issues could arise in any dataset, not just solubility datasets. Since duplicated data have the same activity values but only minor differences in structures (e.g., different tautomeric forms), their presence could contribute to the overfitting of models. We developed models using the "Cure" set of 81935 records from the Meng et al study,[10] which contained >21k duplicated compounds. For cross-validation, the data were split by records and thus the same molecule could be part of training and validation set simultaneously. The presence of duplicates significantly decreased the RMSE of all methods and strongly overfitted performances were obtained.

Table 6. RMSE calculated for kinetic solubility with and without duplicated measurements

| Dataset | records | ChemProp | AttFP | Transformer CNN |
|---|---|---|---|---|
| Clean without duplicates, see Table 1 | 60392 | 0.408 ± 0.002 | 0.443 ± 0.002 | 0.410 ± 0.002 |
| "Cure" set with duplicates from ref[10] | 81935 | 0.356 ± 0.001 | 0.423 ± 0.001 | 0.323 ± 0.002 |

Although the authors made all their scripts publicly available as open source, their re-use is challenging, as considerable time must be spent adapting them (e.g, scripts are linked to the directory structure of one of the authors; limited documentation etc.) as well as the extremely high computational costs associated with performing these analyses. Unfortunately, the authors did not deposit their optimized hyperparameters, developed models, and calculated values, making the reproduction of their results extremely difficult, if not impossible. Thus we recommend that in the future, intermediate logs of results should be reported in addition to scripts, particularly for calculations that require extensive computational power to reproduce final results. The latter aspect will become more critical with the increase of computational resources required to repeat calculations.

# Data Availability

The code used for model development can be found at https://github.com/openochem. This study was carried out using publicly available data from the GitHub repository at https://github.com/Mengjintao/SolCuration. The models and datasets generated during and/or analyzed during the study are exemplified at https://solub.ochem.eu and instructions to reproduce the models are at https://github.com/openochem/openochem/tree/main/solub.

# Funding

This study has received partial funding from the European Union's Horizon 2020 research and innovation program under the Marie Sklodowska-Curie Actions Innovative Training Network European Industrial Doctorate grant agreement "Advanced machine learning for Innovative Drug Discovery (AIDD)" No. 956832.

# Acknowledgement

G.G wants to thank Dr. Valery Normand and Pr. Pierre Vandergheynst for their fruitful and constructive comments during article preparation as well as Dr. Martin Šícho for his contribution to openOCHEM. The authors thanks Dr. Katya Ahmad for her comments and suggestions.

# Author Contributions

Conceptualization: I.V.T, Rv.D, G.G.; methodology: I.V.T., G.G.; software: I.V.T., G.G.; calculations: I.V.T., analysis: I.V.T, Rv.D, G.G.; writing: I.V.T, Rv.D, G.G.

# Competing Interests and Consent for publication

The authors declare that they have no competing interests. All authors have read and agreed to the published version of the manuscript.

doi:10.1093/oso/9780199631469.002.0001.